\documentclass[10pt,twocolumn,letterpaper]{article}

\usepackage[pagenumbers]{iccv} 

\definecolor{iccvblue}{rgb}{0.21,0.49,0.74}
\usepackage[pagebackref,breaklinks,colorlinks,allcolors=iccvblue]{hyperref}
\def\paperID{13550} 
\def\confName{ICCV}
\def\confYear{2025}

\title{FEDS: Feature and Entropy-Based Distillation Strategy for Efficient Learned Image Compression}

\author{
Haisheng Fu\textsuperscript{1} \quad
Jie Liang\textsuperscript{1} \quad
Zhenman Fang\textsuperscript{1} \quad
Jingning Han\textsuperscript{2} \\
\textsuperscript{1}Simon Fraser University \quad
\textsuperscript{2}Google\\
{\tt\small }}

\begin{document}
\maketitle

\begin{abstract}
Learned image compression (LIC) methods have recently outperformed traditional codecs such as VVC in rate-distortion performance. However, their large models and high computational costs have limited their practical adoption.  In this paper, we first construct a high-capacity teacher model by integrating Swin-Transformer V2-based attention modules, additional residual blocks, and expanded latent channels, thus achieving enhanced compression performance. Building on this foundation, we propose a \underline{F}eature and \underline{E}ntropy-based \underline{D}istillation \underline{S}trategy (\textbf{FEDS}) that transfers key knowledge from the teacher to a lightweight student model. Specifically, we align intermediate feature representations and emphasize the most informative latent channels through an entropy-based loss. A staged training scheme refines this transfer in three phases: feature alignment, channel-level distillation, and final fine-tuning. Our student model nearly matches the teacher across Kodak (1.24\% BD-Rate increase), Tecnick (1.17\%), and CLIC (0.55\%) while cutting parameters by about 63\% and accelerating encoding/decoding by around 73\%. Moreover, ablation studies indicate that FEDS generalizes effectively to transformer-based networks. The experimental results demonstrate our approach strikes a compelling balance among compression performance, speed, and model parameters, making it well-suited for real-time or resource-limited scenarios.
\end{abstract}

\section{Introduction} 
\label{sec}

The rapid advancement of learned image compression (LIC) methods has led to performance that surpasses traditional codecs like JPEG~\cite{JPEG}, JPEG2000~\cite{JPEG2000}, BPG~\cite{BPG}, and even the latest H.266/VVC~\cite{VVC} in terms of rate-distortion efficiency. By replacing linear transforms such as the discrete cosine transform (DCT) and discrete wavelet transform (DWT) with deep neural networks, LIC methods can capture complex image structures more effectively, leading to more compact and efficient latent representations.

Early Learned Image Compression (LIC) architectures predominantly utilized Convolutional Neural Networks (CNNs)~\cite{Variational,Joint,GLLMM,Lee_2020,Lee_2021,He_2021_CVPR,He_2022_CVPR,jiang2023mlic}. More recently, transformer-based methods~\cite{Liu_2023_CVPR,zhu2022transformerbased,Qian2022_ICLR} have been introduced, achieving impressive coding performance. However, these transformer-based models often require greater computational resources and are more challenging to train. To further enhance rate-distortion performance without excessively increasing complexity, various modules have been integrated into LIC architectures, including residual blocks~\cite{cheng2020,GLLMM}, attention mechanisms~\cite{cheng2020,Liu_2023_CVPR}, wavelet transform blocks~\cite{Fu_EECV}, and invertible neural networks~\cite{xie2021enhanced}. Despite these advancements, achieving an optimal balance between compression efficiency and computational complexity remains a significant challenge in the field.
\begin{figure*}[!t]
	\centering
		\includegraphics[scale=0.3]{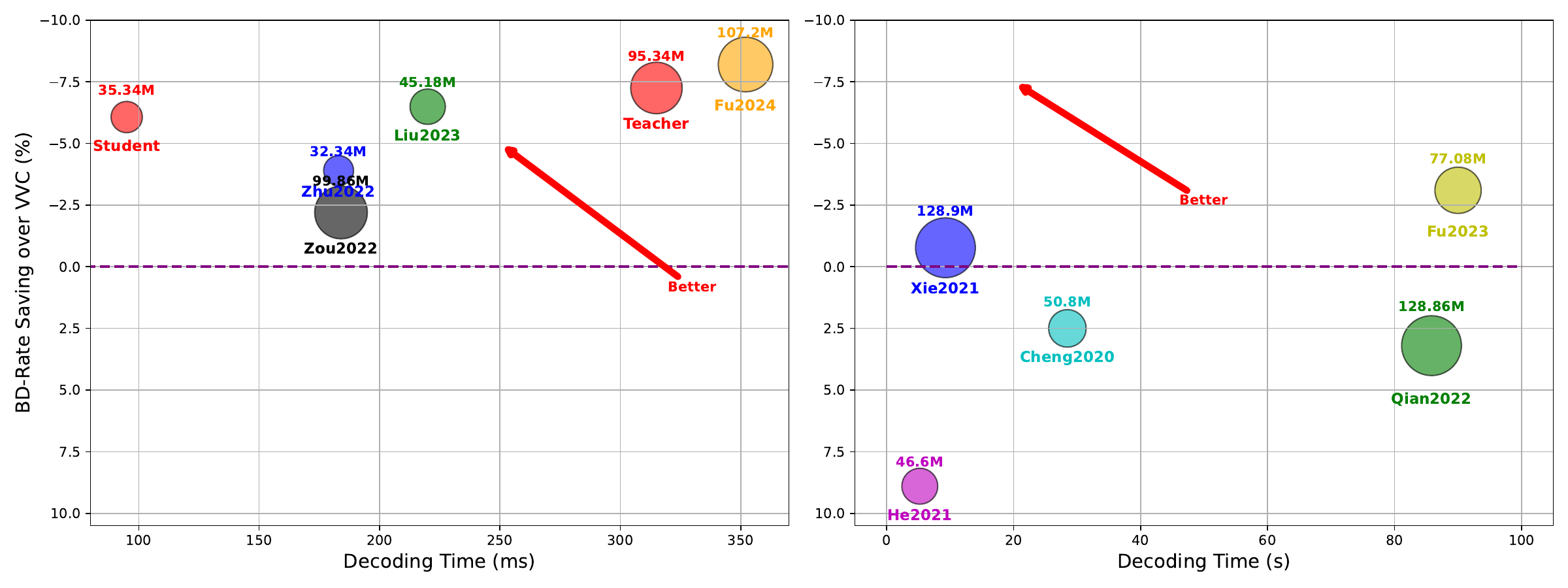}
	\caption{The decoding time and BD-Rate saving over H.266/VVC of different methods for the Kodak dataset. \textbf{Our method achieves the best trade-off among the three metrics}. The upper-left corner has better results. Note that the unit of the left subfigure is \textbf{milliseconds}, while the unit of the right subfigure is \textbf{seconds}. Additionally, the area of each circle represents the model size in terms of the number of parameters.}
	\label{comp_time_bdrate}
\end{figure*}
Adaptive entropy modeling is another key component in improving LIC performance~\cite{Variational, Joint, cheng2020}. However, serial adaptive context models along the spatial dimension disrupt parallelism and significantly slow down decoding. To mitigate this, He et al.~\cite{He_2021_CVPR} proposed checkerboard convolution for parallel entropy modeling, while Minnen et al.~\cite{channel} introduced a channel-based context model. Despite these advancements, achieving significant bit-rate reductions with parallel entropy models remains challenging, as they can still hinder inference speed.

Although these methods have improved rate-distortion performance, many suffer from high model complexity and long processing times, limiting their practical deployment. Balancing compression performance with computational efficiency remains a significant challenge in LIC.

In this paper, we address this challenge by proposing a knowledge distillation framework that effectively reduces model size and computational complexity while maintaining high compression performance. Our contributions are as follows:

\begin{itemize}

\item We build a powerful teacher network by integrating Swin-Transformer V2-based attention modules into the transformation network, incorporating additional residual blocks, and expanding the latent channel dimensionality. These components boost rate-distortion performance and form a robust basis for subsequent knowledge distillation.

\item We propose a dual-stage distillation mechanism: feature-level alignment to preserve intermediate representations and an entropy-based loss to emphasize highly informative latent channels. This enables a lightweight student network to inherit most of the teacher’s coding performance while substantially reducing complexity.

\item We adopt a three-phase training strategy for the student network: (i) feature alignment, (ii) crucial-channel distillation in the latent space, and (iii) fine-tuning. This approach ensures effective knowledge transfer and minimizes performance loss relative to the teacher.

\end{itemize}

As a result (as shown in Fig.~\ref{comp_time_bdrate}), our student network achieves nearly the same performance as the teacher network, with only a 1.24\% increase in BD-Rate, while reducing parameters by approximately 67\% and accelerating encoding and decoding times by around 2.9 times. Compared to recent state-of-the-art methods, our student network provides a better trade-off between coding performance and efficiency on the Kodak dataset. In addition to the Kodak dataset, in Sec. \ref{sec_results}, we provide more comparative results including the Tecnick and CLIC datasets to further demonstrate the advantages of our approach.

\begin{figure}[!t] 
\centering 
\includegraphics[scale=0.30]{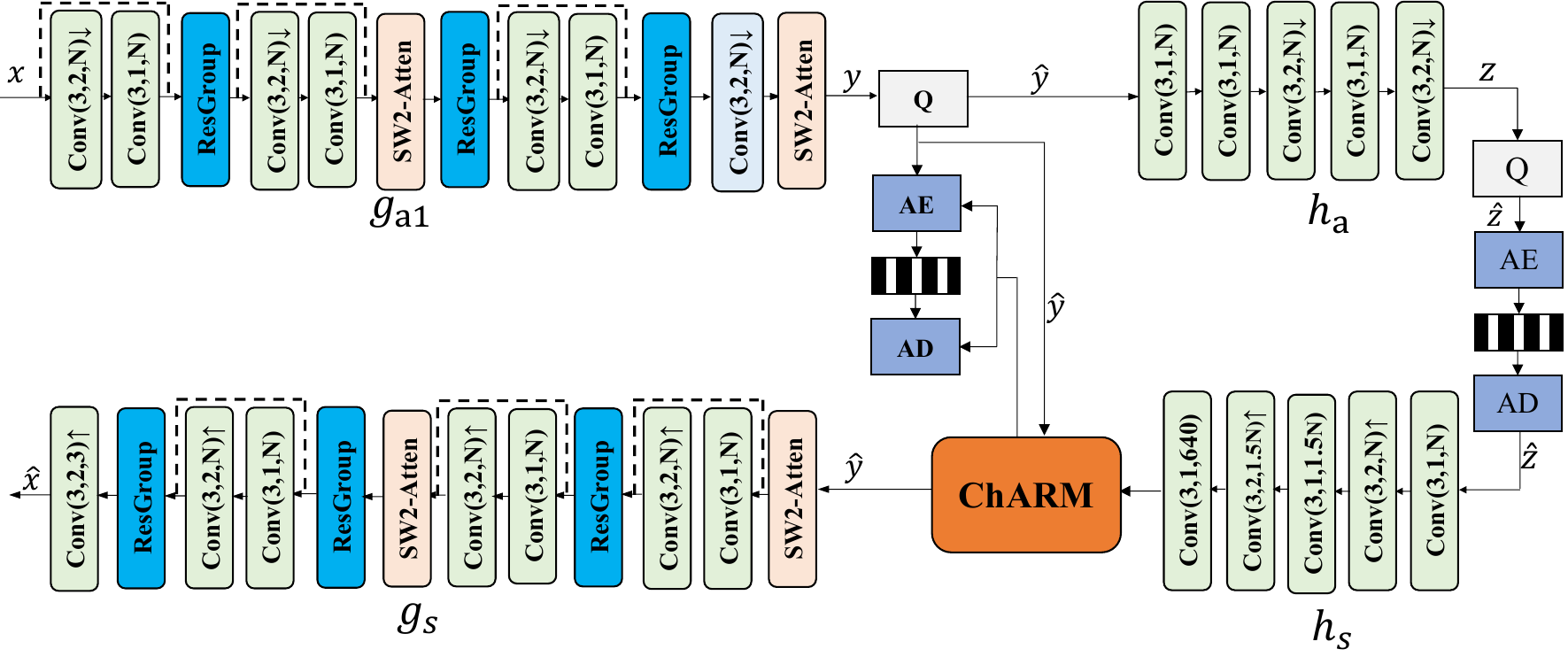} 
\caption{The overall architecture of the proposed \textbf{teacher network}. $Q$ denote quantization module. The symbols $\uparrow$ and $\downarrow$ indicate up-sampling and down-sampling operations, while $3 \times 3$ refers to the convolutional kernel size. $AE$ and $AD$ represent the arithmetic encoder and decoder. The dotted lines illustrate shortcut connections with modified tensor dimensions. $ChARM$ represents he channel-wise auto-regressive entropy model.} 
\label{networkstructure} 
\end{figure}

\section{Related Work} \label{sec_related_work}

\textbf{Learned Image Compression.} Learned image compression methods have made significant strides, outperforming traditional codecs like VVC in rate-distortion performance. The foundational work by Ballé et al.~\cite{Variational} introduced the first end-to-end CNN-based LIC model. Subsequent works incorporated variational autoencoders (VAEs) with hyperpriors~\cite{Variational} to capture spatial dependencies and improve compression efficiency. Minnen et al.~\cite{Joint} proposed a local context model for adaptive auto-regressive entropy modeling, further enhancing compression rates.

Advanced auto-regressive entropy models, such as Gaussian Mixture Models (GMMs)~\cite{cheng2020} and Gaussian-Laplacian-Logistic Mixture Models (GLLMM)~\cite{GLLMM}, have been introduced to estimate more precise probability distributions of latent representations. However, these models are computationally intensive and difficult to accelerate using GPUs due to their sequential nature. To improve computational efficiency, parallelizable entropy models like the checkerboard context model~\cite{He_2021_CVPR} and the channel-wise auto-regressive entropy model (ChARM)~\cite{channel} have been proposed, balancing performance and speed.

Beyond entropy modeling, various architectural enhancements have been explored to improve LIC performance. These include residual networks~\cite{cheng2020, GLLMM}, invertible neural networks~\cite{xie2021enhanced}, octave convolution modules~\cite{octave}, and the integration of transformer modules~\cite{Qian2022_ICLR, zhu2022transformerbased, Liu_2023_CVPR}. While transformer-based methods have achieved impressive results, they are often computationally demanding and require substantial GPU memory, with complexity growing quadratically with input size.

\textbf{Knowledge Distillation.} Knowledge distillation, introduced by Hinton et al.~\cite{hinton2015distilling}, involves training a lightweight student network to mimic the outputs of a larger teacher model. It has been widely applied in computer vision tasks~\cite{Chen_Distillation, Yim_2017_CVPR, gu2022openvocabulary, Li_KD_2022, Yu_KD_2023}. In object detection, Yang et al.~\cite{Yang_detection_KD} proposed Focal and Global Distillation (FGD) to guide student detectors. For image classification, various distillation techniques~\cite{Tung_KD_2019, Heo_KD_2019} have yielded impressive results.

In the context of LIC, knowledge distillation has been less explored. The work in~\cite{GAN_distillation_image_compression} applied distillation to focus on visual performance at low bit rates using GANs but lacked a hypernetwork and did not leverage intermediate features. Fu et al.~\cite{Fu_KD_2024} introduced a distillation method considering both the output and the probability distribution of latent representations, reducing decoder complexity. However, this method did not utilize intermediate feature knowledge and shared the same encoder for both teacher and student networks, complicating optimization and not addressing decoding complexity reduction.

\textbf{Attention Modules.} Attention mechanisms help models focus on important regions, enhancing detail extraction and improving rate-distortion performance. Non-local attention modules~\cite{NonLocal} have been integrated into LIC architectures but are computationally intensive. Simplified local attention modules~\cite{cheng2020} and window-based attention modules~\cite{Liu_2023_CVPR} have been proposed to accelerate computation while retaining performance benefits.

In our work, we build upon these advances by integrating Swin-Transformer V2-based attention modules into our teacher network to enhance performance, and then effectively distilling this knowledge into a lightweight student network.

\section{The Proposed Image Compression Framework} 
\label{sec_method}

In this section, we first present the detailed architecture of the teacher network, highlighting the specific components that enhance its performance. We then describe how the student network is derived from the teacher network by simplifying certain modules, and we elaborate on the knowledge distillation process that transfers the teacher's knowledge to the student network. Finally, we address the training strategy employed for both networks.

\subsection{Teacher Network Architecture} 
\label{sec_teacher_network}

The architecture of the proposed teacher network is illustrated in Fig.~\ref{networkstructure}. The input image $\mathbf{x} \in \mathbb{R}^{H \times W \times 3}$ is processed by the main encoder network $g_{a}$, which extracts a compact latent representation $\mathbf{y}$. To enhance the capacity of the teacher network, we introduce the following key components:

\textbf{Swin-Transformer V2 Attention Modules:} We integrate two attention modules based on Swin-Transformer V2~\cite{swinTransformerV2} into the transformation network, and we will describe their architectures in Sec. \ref{sec_SW2V2}. These modules are placed after the first and second down-sampling layers, allowing the network to capture both local and global dependencies effectively.

\textbf{Increased Residual Blocks:} We add three groups of residual blocks, each containing six basic residual blocks~\cite{resblock}, to deepen the network and enhance its representation capacity.  

\textbf{Expanded Latent Representation Channels:} We increase the number of channels in the latent representation $\mathbf{y}$ from the standard 192 to 400, providing a richer feature space for encoding image information. 

The main encoder $g_{a}$ consists of four stages of down-sampling, each reducing the spatial dimensions by a factor of 2. The Swin-Transformer V2 attention modules and residual blocks are interleaved within these stages, as shown in Fig.~\ref{networkstructure}. The main decoder $g_{s}$ mirrors the encoder architecture, using up-sampling layers to restore the spatial dimensions.

For entropy coding, we employ a channel-wise auto-regressive entropy model (ChARM)~\cite{Ma_IEEE_2020,channel}, which efficiently models the distribution of $\mathbf{y}$ by exploiting the correlations across channels. This model allows for parallel encoding and decoding, significantly accelerating the compression process. We will describe the detailed content in the appendix (Sec. \ref{ChARM}).

We designed our own teacher network based on a CNN architecture to establish a flexible and representative foundation for our proposed compression framework. By constructing a teacher network that incorporates typical components of state-of-the-art compression models, we can better understand the factors affecting compression performance. Demonstrating that our distillation and training strategies are effective on this prototypical network suggests that they can be applied broadly to other models in the field, promoting a general and adaptable compression framework. We will also demonstrate that our proposed modules can be used in transformer-based methods in later ablation experiments.

\subsection{Student Network Architecture} 
\label{sec_student_network}

To obtain a lightweight student network, we simplify the teacher network by reducing model complexity while aiming to maintain performance. The modifications are as follows:

\textbf{Removal of Attention Modules:} We eliminate the two Swin-Transformer V2 attention modules from the transformation network, reducing computational overhead. 

\textbf{Reduced Residual Blocks:} We decrease the number of residual blocks in each group from six to one, simplifying the network architecture. 

\textbf{Reduced Latent Representation Channels:} We reduce the number of channels in the latent representation $\mathbf{y}$ from 400 to 160, decreasing the model size and computation. 

The student network retains the overall structure of the teacher network but with these simplifications. This results in a model that is significantly smaller and faster, making it more suitable for practical deployment.

\subsection{Knowledge Distillation Framework} \label{sec_distillation}

\begin{figure}[!thp] \centering \includegraphics[scale=0.44]{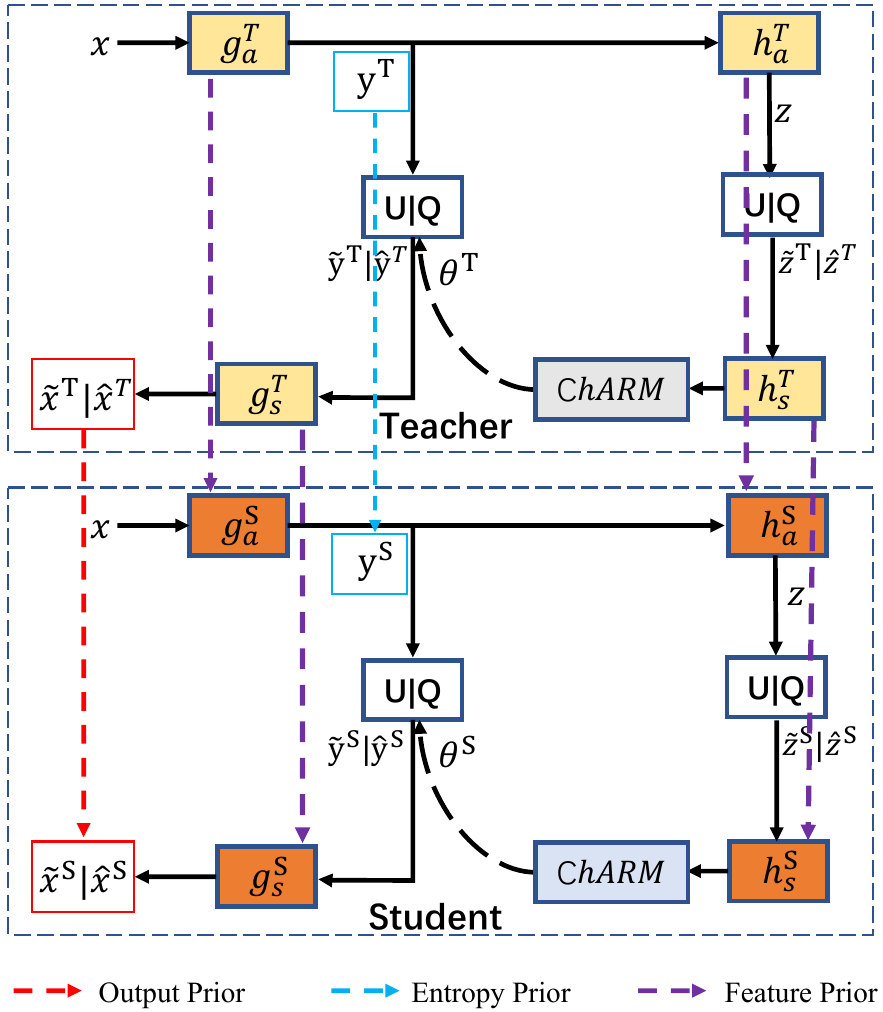} \caption{The knowledge distillation framework between the teacher and student networks.} \label{distillation_framework_model} \end{figure}

In this section, we propose a knowledge distillation framework (as shown in Figure~\ref{distillation_framework_model}) to reduce the complexity of the neural image compression model while preserving high performance. Our approach involves training a high-capacity teacher network and transferring its knowledge to a more efficient student network. The entire training process consists of three stages.

\textbf{First}, we train the teacher network, which has a larger number of parameters and higher computational complexity, using the standard rate-distortion optimization framework. The loss function for the teacher network is:

\begin{equation}\label{teacher_loss} 
\begin{aligned} L_{T} &= D(x,\hat{x}^T) +\lambda (R_y^T + R_z^T),\\
\ R_y^T &= E [-\log_{2}(P_{\hat{y}^T|\hat{z}^T}(\hat{y}^T|\hat{z}^T))], \\
\ R_z^T &= E [-\log_{2}(P_{\hat{z}^T}(\hat{z}^T))], \end{aligned} 
\end{equation}

where $D(x, \hat{x}^T)$ is the distortion between the original image $x$ and the reconstructed image $\hat{x}^T$ from the teacher network, typically measured by Mean Squared Error (MSE) or Multi-Scale Structural Similarity (MS-SSIM). $R_y^T$ and $R_z^T$ represent the estimated bit rates of the quantized latent representations $\hat{y}^T$ and $\hat{z}^T$, respectively. $\lambda$ is the Lagrange multiplier that controls the trade-off between rate and distortion.

\textbf{Second}, we design a student network with reduced complexity by decreasing the number of channels in the latent representation (e.g., reducing $M$ from 320 to 160) and simplifying the network architecture, such as removing attention modules and reducing the number of residual blocks. To effectively transfer knowledge from the teacher network to the student network, we jointly train the student network using knowledge distillation. Our knowledge distillation framework is illustrated in Fig.~\ref{distillation_framework_model}, where superscripts $T$ and $S$ denote the teacher and student networks, respectively.

The loss function for training the student network with knowledge distillation is defined as:

\begin{equation}
\label{total_loss} 
\begin{aligned} L_{S} &= D(x,\hat{x}^S)+\lambda (R_y^S + R_z^S) + L_{\text{KD}}, \ 
\end{aligned} 
\end{equation}

where $\hat{x}^S$ is the reconstructed image from the student network, and $R_y^S$, $R_z^S$ are the estimated bit rates of the student's latent representations $\hat{y}^S$ and $\hat{z}^S$. $L_{\text{KD}}$ is the knowledge distillation loss, which consists of three components:

\begin{equation} L_{\text{KD}} = \alpha L_{\text{output}} + \beta L_{\text{feature}} + \gamma L_{\text{latent}}, \end{equation}

where $\alpha$, $\beta$, and $\gamma$ are hyperparameters that control the relative importance of each distillation component.

The \textbf{output distillation loss} $L_{\text{output}}$ encourages the student network to produce reconstructed images similar to those of the teacher network:

\begin{equation} L_{\text{output}} = d(\hat{x}^T, \hat{x}^S), \end{equation}

where $d(\cdot, \cdot)$ is a distortion measure, such as MSE.

The \textbf{feature distillation loss} $L_{\text{feature}}$ aims to align the intermediate feature representations between the teacher and student networks. By doing so, we encourage the student network to learn similar hierarchical abstractions as the teacher network, which can enhance its performance despite reduced complexity.

We compute $L_{\text{feature}}$ by measuring the discrepancy between corresponding feature maps extracted from both networks:

\begin{equation} L_{\text{feature}} = \frac{1}{N} \sum_{i=1}^{N} d\left( f_i^T, f_i^S \right), \end{equation}

where $f_i^T$ and $f_i^S$ are the feature maps from the $i$-th layer of the teacher and student networks, respectively, and $N$ is the total number of selected layers. The function $d(\cdot, \cdot)$ denotes a distortion measure, such as Mean Squared Error (MSE).

To ensure compatibility between the feature maps, we select layers where the spatial dimensions and number of channels match or can be appropriately adjusted. In cases where the dimensions differ due to architectural changes (e.g., different numbers of channels), we either exclude those layers from the loss computation or apply a simple transformation to align them.

By minimizing $L_{\text{feature}}$, the student network is guided to replicate the teacher's internal representations. This process facilitates the transfer of knowledge at multiple abstraction levels, enabling the student network to capture essential features necessary for high-performance image compression.

The \textbf{Entropy-Based Distillation loss} $L_{\text{latent}}$ focuses on transferring the most informative parts of the teacher's latent representation to the student network. Specifically, we select the top $C_s$ channels from the teacher's latent representation $\hat{y}^T$ based on their importance, where $C_s$ is the number of channels in the student's latent representation $\hat{y}^S$. The importance of each channel is measured by the entropy of its corresponding probability distribution in the entropy model.

To compute the importance of each channel in $\hat{y}^T$, we calculate the channel-wise entropies using the learned entropy model. For each channel $c$ in $\hat{y}^T$, the entropy is computed as:

\begin{equation} 
H_c = -E_{\hat{y}^T_c} \left[ \log_{2} P_{\hat{y}^T_c | \hat{z}^T} (\hat{y}^T_c | \hat{z}^T) \right], 
\end{equation}

where $\hat{y}^T_c$ represents the $c$-th channel of the teacher's latent representation, and $P_{\hat{y}^T_c | \hat{z}^T}$ is the conditional probability distribution estimated by the entropy model. The expectation is taken over the spatial dimensions of the feature maps.

We then rank all channels in $\hat{y}^T$ based on their entropies $H_c$ in descending order. The top $C_s$ channels with the highest entropies are selected as they carry the most information content. This selection process can be formally expressed as:

\begin{equation} 
{ c_1, c_2, \dots, c_{C_s} } = \operatorname{TopK} ( { H_1, H_2, \dots, H_{C_t} }, C_s ), 
\end{equation}

where $C_t$ is the total number of channels in $\hat{y}^T$, and $\operatorname{TopK}$ returns the indices of the top $C_s$ channels with the highest entropies.

The latent space distillation loss is then defined as:

\begin{equation} 
L_{\text{latent}} = d\left( \hat{y}^T_{\text{selected}}, \hat{y}^S \right), 
\end{equation}

where $\hat{y}^T_{\text{selected}}$ consists of the selected top $C_s$ channels from the teacher's latent representation, and $\hat{y}^S$ is the student's latent representation. By minimizing $L_{\text{latent}}$, we encourage the student network to capture the most informative features learned by the teacher network.

By incorporating these distillation losses, we encourage the student network to mimic the behavior of the teacher network at multiple levels, leading to improved performance.

\textbf{Third}, after training the student network with knowledge distillation, we further fine-tune the student network independently without the teacher network. This stage allows the student network to adapt its parameters fully and potentially improve performance by focusing on its own optimization objectives.

\begin{figure*} 
\begin{subfigure}{0.45\linewidth} 
\centering \includegraphics[width=\columnwidth]{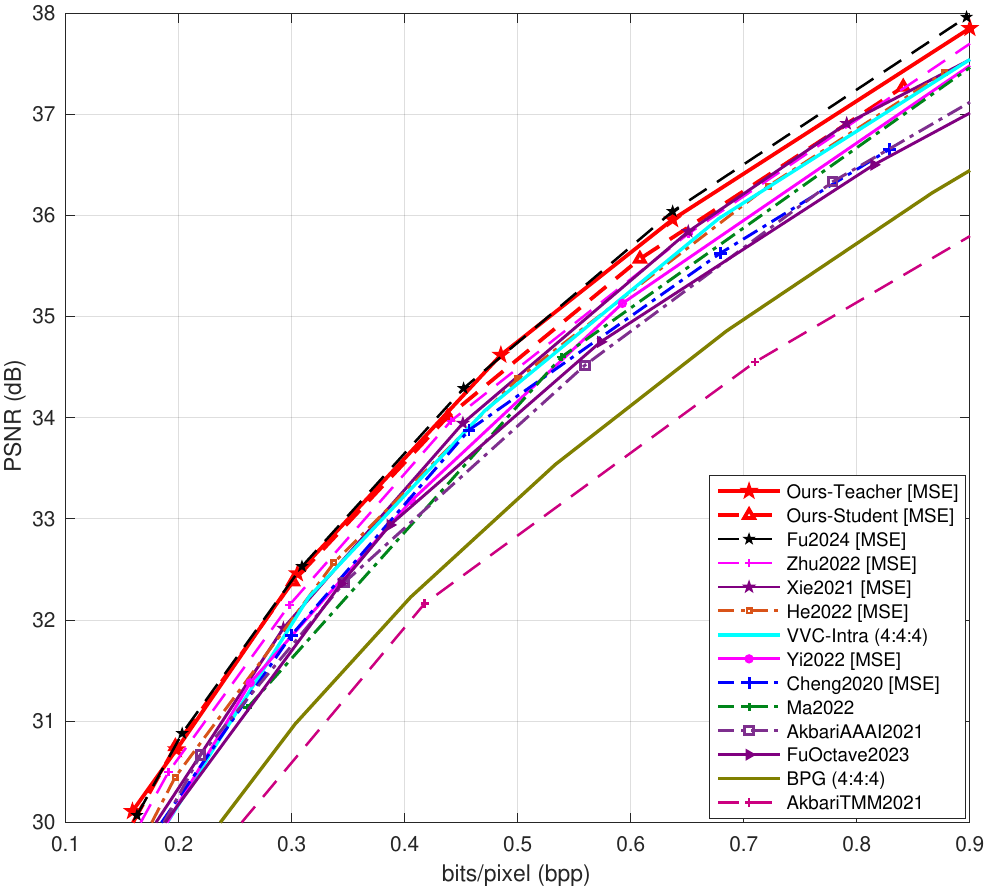} 
\caption{} \label{kodak_PSNR} \end{subfigure} 
\hfill 
\begin{subfigure}{0.45\linewidth} 
\centering 
\includegraphics[width=\columnwidth]{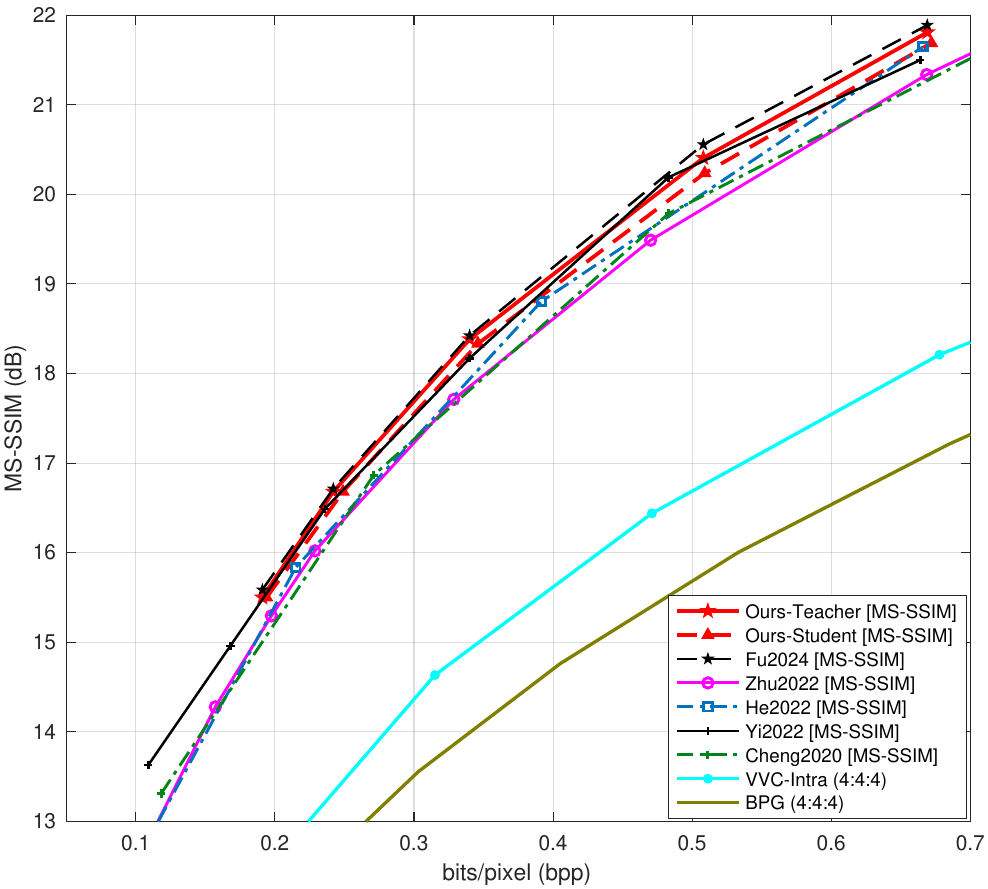} 
\caption{} 
\label{kodak_SSIM} 
\end{subfigure} 
\caption{The R-D curves of different methods in terms of PSNR and MS-SSIM on the Kodak dataset \cite{Kodak}.} 
\label{fig:kodok} 
\end{figure*}

\begin{figure*} 
\begin{subfigure}{0.45\linewidth} 
\centering \includegraphics[width=\columnwidth]{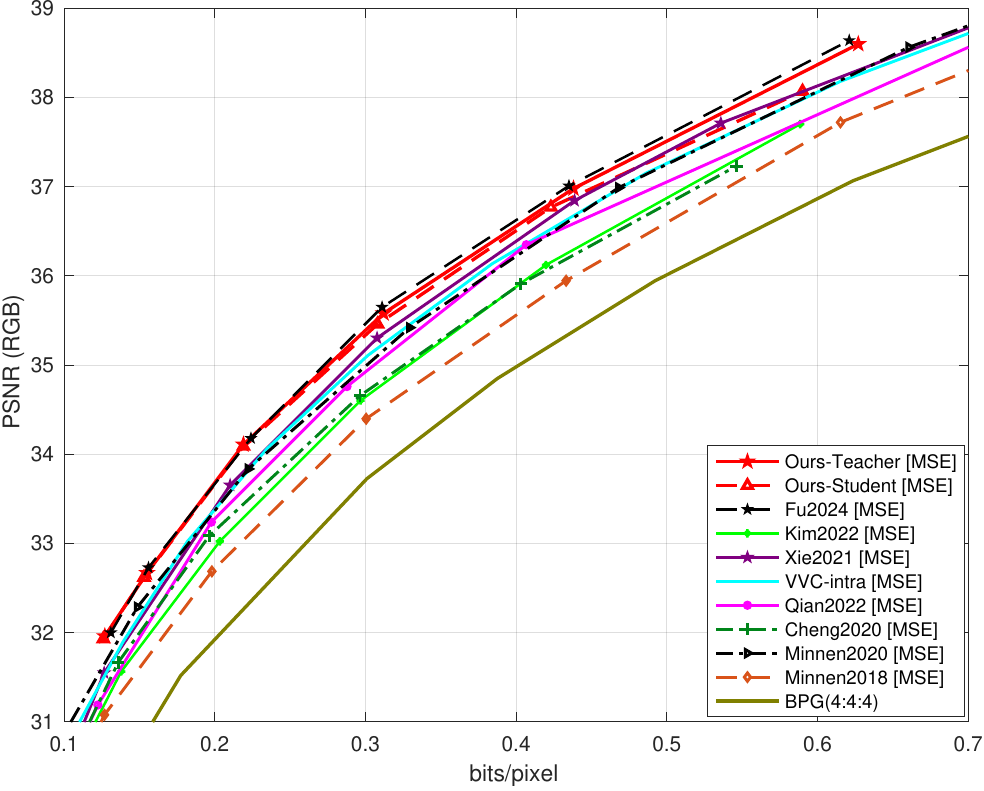} \caption{} 
\label{fig:Tecnick} 
\end{subfigure} 
\hfill 
\begin{subfigure}{0.45\linewidth} 
\centering 
\includegraphics[width=\columnwidth]{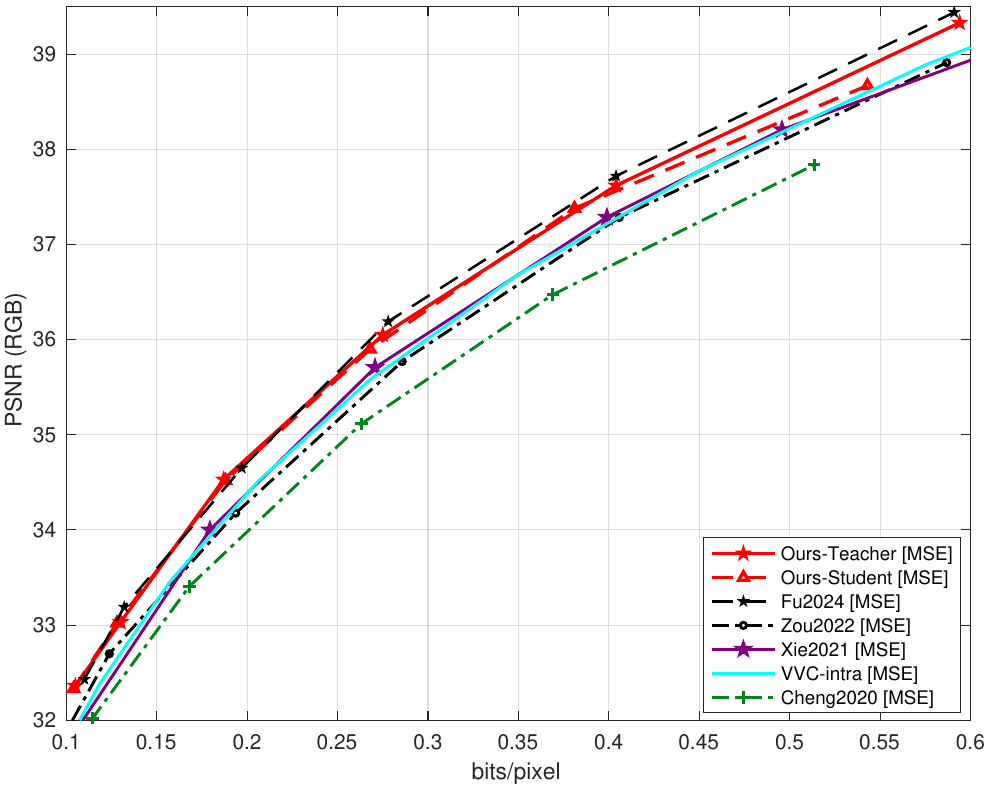} \caption{} 
\label{fig:CLIC} 
\end{subfigure} 
\caption{The R-D curves of different methods in terms of PSNR on the Tecnick-100  \cite{Tecnick} and CLIC-2021-test \cite{CLIC_test_2021} datasets.} 
\label{fig} 
\end{figure*}

\section{Experimental Results}
\label{sec_results}

In this section, we evaluate the performance of the proposed teacher and student networks against several state-of-the-art Learned Image Compression (LIC) methods and traditional image codecs, using both Peak Signal-to-Noise Ratio (PSNR) and Multi-Scale Structural Similarity Index (MS-SSIM) metrics. The LIC methods considered include Fu2024~\cite{Fu_EECV}, Liu2023~\cite{Liu_2023_CVPR}, Fu2023~\cite{GLLMM}, FuOctave2023~\cite{Fu_octave}, Kim2022~\cite{Kim_cvpr2022}, Zhu2022~\cite{zhu2022transformerbased}, Qian2022~\cite{Qian2022_ICLR}, He2022~\cite{He_2022_CVPR}, He2021~\cite{He_2021_CVPR}, Xie2021~\cite{xie2021enhanced}, AkbariAAAI2021~\cite{Mahammand_AAAI}, AkbariTMM2021~\cite{Mahammand_media}, Cheng2020~\cite{cheng2020}, Minnen2020~\cite{channel}, and Minnen2018~\cite{Joint}. For traditional codecs, we include H.266/VVC Intra (4:4:4) and H.265/BPG Intra (4:4:4).

Our method is evaluated on three widely-used test datasets: the Kodak PhotoCD test set~\cite{Kodak}, which contains 24 images at resolutions of $768 \times 512$ or $512 \times 768$; the Tecnick-40 test set~\cite{Tecnick}, comprising 40 images at a resolution of $1200 \times 1200$; and the CLIC 2021 test set~\cite{CLIC_test_2021}, which includes 60 images with resolutions ranging from $751 \times 500$ to $2048 \times 2048$.

To ensure a fair comparison, we adapted the Cheng2020~\cite{cheng2020} method by increasing the number of filters $N$ from 192 to 256 for higher bit-rate scenarios, leading to improved performance over the originally reported results in~\cite{cheng2020}. Results for other methods were obtained from their open-source implementations or directly from their respective publications.

We present the results for both the teacher and student networks. The teacher network employs the same architecture depicted in Fig.~\ref{networkstructure}. In the student network, we simplify the architecture by removing the attention and residual modules in the main encoder $g_{a}$ and main decoder $g_{s}$ and by reducing the latent representation size $M$ from 400 to 160.

\subsection{R-D Performance}

Figure~\ref{fig:kodok} presents the average Rate-Distortion (R-D) curves of various methods on the Kodak dataset, evaluated using PSNR and MS-SSIM metrics. Among PSNR-optimized methods, Fu2024(MSE)~\cite{Fu_KD_2024} demonstrates the best performance over a wide range of bit rates, surpassing H.266/VVC and other learned methods across all bit rates. Our proposed \textbf{Teacher} method matches the performance of Fu2024(MSE) when the bit rate is below 0.6 bpp and consistently outperforms Zhu2022~\cite{zhu2022transformerbased} by approximately 0.2dB across all bit rates. At higher bit rates, our results are slightly worse than Fu2024(MSE), which can be attributed to Fu2024 quadrupling the number of channels $M$ via wavelet transform. Regarding the MS-SSIM metric in Figure~\ref{kodak_SSIM}, Fu2024(MS-SSIM) also achieves the highest performance among the compared methods. Our \textbf{Teacher} method matches the performance of Fu2024(MS-SSIM) at low bit rates and is slightly worse at high bit rates. Our \textbf{Student} method nearly achieves the same performance as the \textbf{Teacher} method, with a slight decrease at high bit rates.

Figure~\ref{fig:Tecnick} shows the PSNR performance of different methods on the Tecnick-40 dataset. Among the PSNR-optimized methods, our \textbf{Teacher} method leads the compared methods. The Fu2024 method surpasses our method only at high bit rates. Both our methods also achieve better performance than H.266/VVC across a wide range of bit rates.

Figure~\ref{fig:CLIC} compares PSNR performance on the CLIC2021 test set. Fu2024~\cite{Fu_EECV} achieves the best results among all compared methods. Our \textbf{Teacher} approach attains significant gains over Xie2021~\cite{xie2021enhanced}, Zou2022~\cite{Zou_2022_CVPR}, and H.266/VVC, with up to 0.5~dB improvement at higher bit rates.

\subsection{Performance and Speed Trade-off}

\begin{table}[!t]
\caption{Comparisons of encoding/decoding times, BD-Rate reductions over VVC, and model parameters on the Kodak, Tecnick, and CLIC test sets.}
\begin{tabular}{ccccc}
\hline
\textbf{Methods} & \textbf{Enc} & \textbf{Dec} & \textbf{BD-Rate}  &\textbf{$\#$Para}\\ 
\hline
\multicolumn{5}{c}{\textbf{Kodak}}\\
\hline
  VVC  & 402.3s & 0.61s & 0.0 & -\\
Cheng2020 \cite{cheng2020}  & 27.6s & 28.8s & 2.6\% & 50.8 M \\
Hu2021 \cite{Hu_2021}  & 32.7s & 77.8s & 11.1\% & 84.6 M \\
He2021 \cite{He_2021_CVPR}  & 20.4s & 5.2s & 8.9\% & 46.6 M \\ 
Xie2021 \cite{xie2021enhanced} & 4.1s & 9.250s & -0.8\% & 128.9 M\\
Zou2022 \cite{Zou_2022_CVPR} & 0.165s & 0.19s & -2.2\% & 99.9 M\\
Zhu2022 \cite{zhu2022transformerbased} & 0.27s & 0.183s & -3.9\% & 32.34 M\\
Liu2023 \cite{Liu_2023_CVPR} & 0.22s & 0.23s & -6.49\% & 45.18 M \\
Fu2024 \cite{Fu_EECV} & 0.352s & 0.388s & -8.2\% & 107.2 M\\
\textbf{Teacher}  & \textbf{0.239s} & \textbf{0.287s} & \textbf{-7.24\%} & \textbf{95.43 M}\\
\textbf{Student}  & \textbf{0.064s} & \textbf{0.096s} & \textbf{-6.0\%} & \textbf{35.34 M}\\ 
\hline
\multicolumn{5}{c}{\textbf{Tecnick}}\\
\hline
VVC  & 700.59s & 1.49s & 0.0 & - \\
Zou2022 \cite{Zou_2022_CVPR} & 0.431s & 0.472s & -2.6\% & 99.9 M\\
Liu2023 \cite{Liu_2023_CVPR} & 0.497s & 0.583s & -8.34\% & 45.18 M\\
Fu2024 \cite{Fu_EECV} & 1.383s & 1.432s & -9.46\% & 107.2 M\\
\textbf{Teacher}  & \textbf{0.576s} & \textbf{0.684s} & \textbf{-8.63\%} & \textbf{95.43 M}\\
\textbf{Student}  & \textbf{0.209s} & \textbf{0.261s} & \textbf{-7.46\%} & \textbf{35.34 M}\\ 
\hline
\multicolumn{5}{c}{\textbf{CLIC}}\\
\hline
VVC  & 949.58s & 1.98s & 0.0 & - \\
Zou2022 \cite{Zou_2022_CVPR} & 0.163s & 0.18s & 0.7855\% & 99.9 M\\
Liu2023 \cite{Liu_2023_CVPR} & 0.791s & 0.895s & -7.68\% & 45.18 M\\
Fu2024 \cite{Fu_EECV} & 2.27s & 2.26s & -9.20\% & 107.2 M\\
\textbf{Teacher}  & \textbf{0.929s} & \textbf{1.08s} & \textbf{-8.18\%} & \textbf{95.43 M}\\
\textbf{Student}  & \textbf{0.324s} & \textbf{0.394s} & \textbf{-7.63\%} & \textbf{35.34 M}\\ 
\hline
\end{tabular}
\label{runing_time}
\end{table}

\begin{table}[h]
    \caption{Comparison of FLOPs and parameters for select methods, evaluated using a 2K image (2560$\times$1440).}
    \begin{tabular}{cccc}
        \toprule
           Liu2023 \cite{Liu_2023_CVPR} & Fu2024 \cite{Fu_EECV} & \textbf{Teacher}  & \textbf{Student}\\
        \midrule
        2.15T & 9.21T & 6.18T & 1.9T \\
        \bottomrule
    \end{tabular}
    \label{tab:flops_comparison}
\end{table}

Table~\ref{runing_time} presents a comprehensive comparison of encoding/decoding times, BD-Rate reductions relative to VVC~\cite{BDRate}, and model sizes across three benchmark datasets: Kodak, Tecnick, and CLIC. Except for VVC—which is evaluated on a CPU-based platform (2.9GHz Intel Xeon Gold 6226R CPU)—all methods are assessed on an NVIDIA Tesla V100 GPU with 16GB memory. For a fair FLOPs comparison (Table~\ref{tab:flops_comparison}), evaluations are conducted using a 2K resolution image (2560$\times$1440).

On the Kodak dataset, Fu2024~\cite{Fu_EECV} achieves the highest BD-Rate reduction of -8.2\%, albeit at the expense of a larger model (107.2M parameters) and longer processing times (0.352s for encoding and 0.388s for decoding). Liu2023~\cite{Liu_2023_CVPR} offers a balanced trade-off with a -6.49\% BD-Rate reduction, 0.22s encoding, 0.23s decoding, and a moderate model size of 45.18M parameters. Our \textbf{Teacher} model attains a -7.24\% BD-Rate reduction with encoding and decoding times of 0.239s and 0.287s, respectively.

In contrast, our proposed \textbf{Student} model, while exhibiting a slightly lower BD-Rate reduction of -6.0\% on Kodak, significantly enhances computational efficiency. It reduces encoding and decoding times to 0.064s and 0.096s, respectively, and employs a compact model of 35.34M parameters. Similar trends are observed on the Tecnick and CLIC datasets, where the Student model achieves BD-Rate reductions of -7.46\% and -7.63\%, respectively, compared to -8.63\% and -8.18\% for the Teacher model.

A detailed comparison reveals that the Student model incurs a performance drop of 1.24 percentage points on Kodak, 1.17 percentage points on Tecnick, and 0.55 percentage points on CLIC relative to the Teacher model. These marginal losses are accompanied by substantial gains in efficiency; the Student model reduces the number of parameters by nearly 63\% (from 95.43M to 35.34M) and significantly lowers decoding times (e.g., from 0.287s to 0.096s on Kodak).

In summary, our Student model achieves an optimal balance among compression efficiency, processing speed, and model compactness, making it highly suitable for real-time and resource-constrained applications.

\subsection{Ablation Studies}
\label{sec:ablation}

\subsubsection{Teacher Network Components}
\label{sec:ablation_teacher}

\begin{figure}[!t]
\centering
\includegraphics[width=0.6\linewidth]{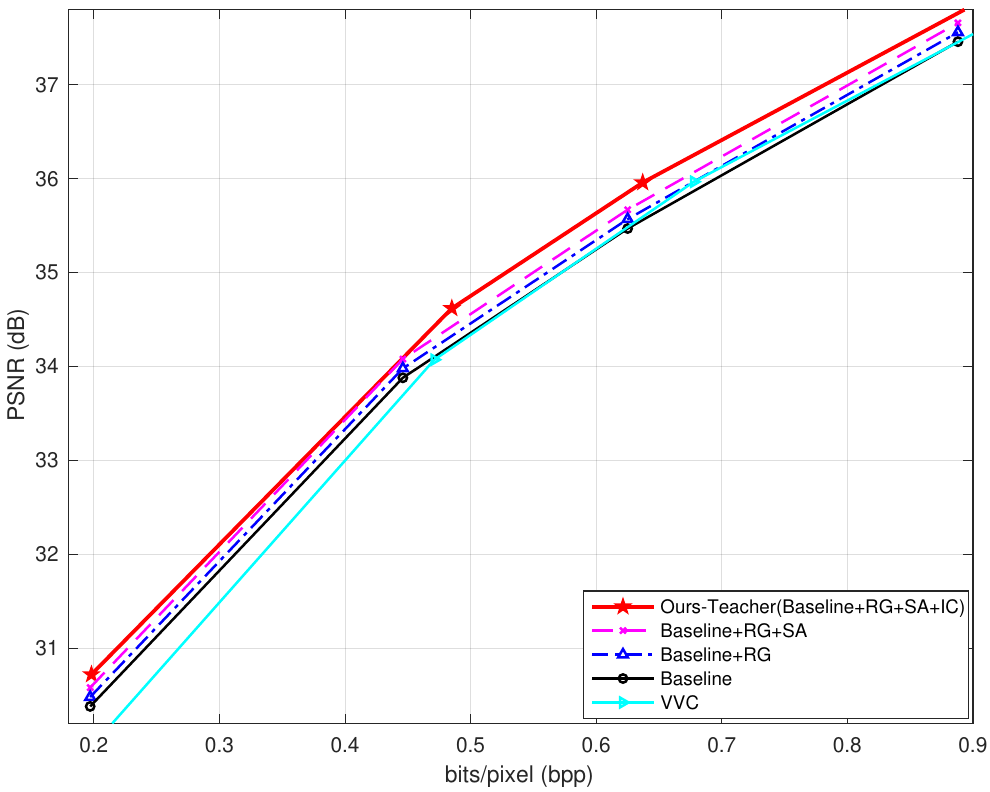}
\caption{Ablation study on the Teacher network, showing the impact of adding Residual Groups (RG), Swin-Transformer V2 Attention modules (SA), and increasing the number of latent channels (IC).}
\label{fig:Teacher_abalation}
\end{figure}

We conducted ablation studies to analyze how each key component affects the Teacher network's performance. All reported results are averaged over the Kodak dataset.

We define a \textbf{Baseline} model by removing the residual groups and the Swin-Transformer V2 attention modules, while setting the number of latent representation channels to 320.

To understand the contribution of each component, we incrementally add:
\begin{itemize}
    \item \textbf{Baseline + RG}: Includes Residual Groups (RG).
    \item \textbf{Baseline + RG + SA}: Further adds Swin-Transformer V2 Attention (SA).
    \item \textbf{Ours Teacher (Baseline + RG + SA + IC)}: Increases the number of latent channels from 320 to 400, denoted as IC (increased channels).
\end{itemize}

As shown in Figure~\ref{fig:Teacher_abalation}, adding Residual Groups (\textbf{Baseline + RG}) improves PSNR by roughly 0.05\,dB. Incorporating Attention modules (\textbf{Baseline + RG + SA}) provides an additional gain of about 0.1\,dB. Increasing the number of channels (\textbf{Baseline + RG + SA + IC}) offers a further 0.06\,dB improvement, especially at higher bit rates. Although augmenting channels can benefit performance, it also increases encoding and decoding times.

\subsubsection{Student Network Distillation Strategies}
\label{sec:ablation_student}

\begin{figure}[!t]
\centering
\includegraphics[width=0.6\linewidth]{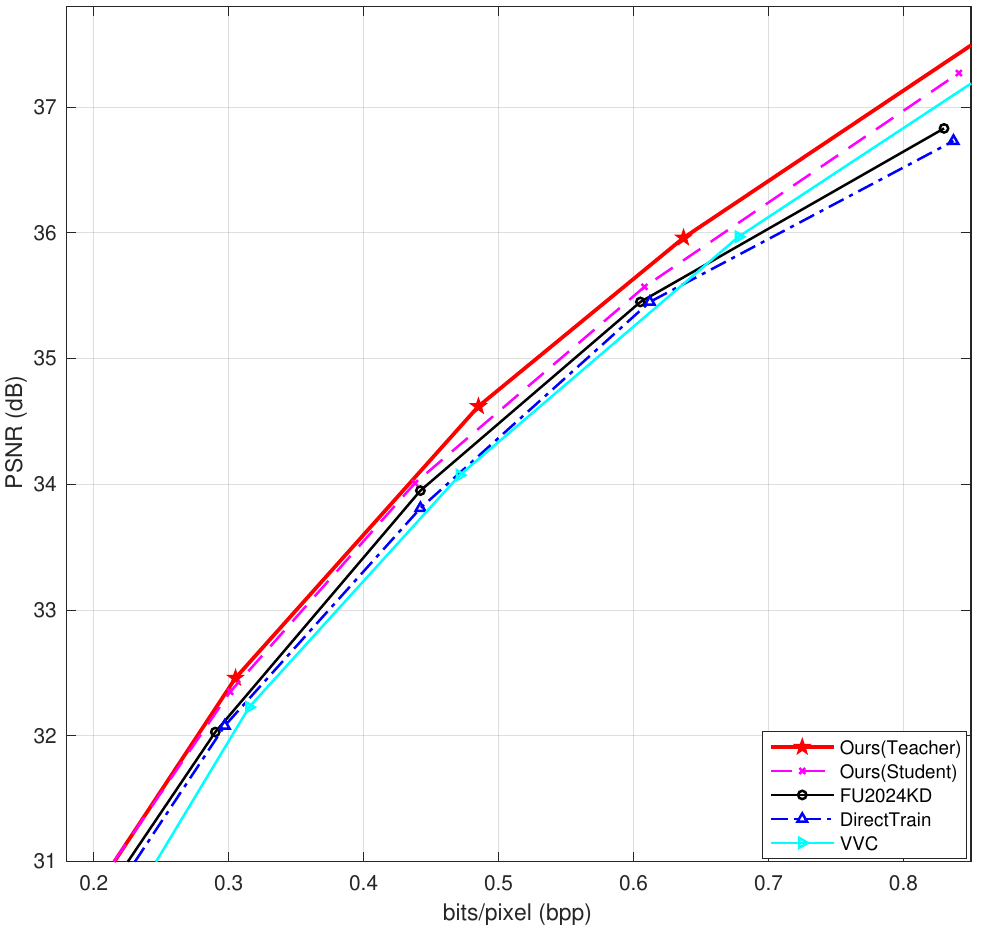}
\caption{Ablation study on the Student network, comparing different training strategies: Direct Training, \textbf{FU2024KD}~\cite{Fu_KD_2024}, and our proposed Distillation method.}
\label{fig:KD_abalation}
\end{figure}

We further evaluated how our proposed knowledge distillation approach influences the Student network’s performance. Figure~\ref{fig:KD_abalation} compares three training strategies:

\begin{itemize}
\item \textbf{Direct Training}: The Student network is trained independently, without any distillation from the Teacher network.
\item \textbf{FU2024KD}~\cite{Fu_KD_2024}: The Student network follows a distillation method introduced by Fu2023, focusing on output and probability distributions.
\item \textbf{Our Distillation}: The Student network employs our proposed feature distillation and entropy-based loss, with a staged training strategy.
\end{itemize}

The experiments reveal that \textbf{Direct Training} suffers noticeable performance drops at both low and high bit rates compared to the Teacher network. While \textbf{FU2024KD} helps recover some performance, it still shows a clear gap. By contrast, our approach effectively leverages essential features from the Teacher network, achieving nearly identical performance at low bit rates and only minor degradation at higher bit rates. Overall, this leads to a BD-Rate increase of merely 1.2\% relative to the Teacher network.

\subsubsection{Distillation Strategies on transformer-based frameworks}
\label{sec:ablation_student}

To further investigate the effectiveness of our proposed approach, we applied it to a transformer-based framework. In this study, we selected a model from the literature as the baseline. First, we increased the number of Swin-Transformer blocks in the baseline from 1 to 3 and then incorporated our Swin-Transformer V2 blocks into the framework. Additionally, we increased the channel dimension from 320 to 400 for the teacher model. For the student model, we removed the added components and set the latent representation channel dimension to 160. The obtained results are shown in Table \ref{liu_cvpr2023}.

\begin{table}[!t]
\caption{Comparisons of encoding/decoding time, BD-Rate reduction over VVC, and model parameters on the Kodak test set.}
\begin{center}
\begin{tabular}{ccccc}
\hline
\textbf{Methods} & \textbf{Enc} & \textbf{Dec} & \textbf{BD-Rate} &\textbf{$\#$Para}\\ 
\hline
\textbf{Liu2023} \cite{Liu_2023_CVPR} &0.22s&0.23s&-6.49\%&45.18 M \\
\textbf{Teacher}  &\textbf{0.239s}& \textbf{0.287s}& \textbf{-9.67\%} &  \textbf{62.46 M}\\
\textbf{Student}  &\textbf{0.124s}& \textbf{0.175s}& \textbf{-8.46\%} &  \textbf{32.34 M}\\ 
   \hline
\end{tabular}

\label{liu_cvpr2023}
\end{center}
\end{table}

Compared to the baseline, our teacher model achieved a 3.18\% reduction in BD-rate, demonstrating the effectiveness of our proposed modules and strategies. This confirms that our approach not only enhances performance in CNN-based models but also improves performance in transformer-based models. Although the student model exhibited a 1.21\% performance drop compared to the teacher model, it still outperformed the baseline with smaller parameters, further validating the effectiveness of our proposed model.

\section{Conclusion}

In this paper, we presented a knowledge distillation framework that transfers key knowledge from a high-capacity teacher network to a lightweight student network for learned image compression. By integrating feature distillation, an entropy-based loss, and a staged training strategy, our student network nearly matches the teacher’s performance on three common datasets while significantly reducing model size and computational load. This demonstrates that knowledge distillation can help advanced compression models become more practical for real-world applications. In future work, we aim to explore more efficient distillation techniques to further narrow the gap between student and teacher networks.

{
    \small
    \bibliographystyle{ieeenat_fullname}

}
\clearpage
\setcounter{page}{1}
\maketitlesupplementary


In this appendix, we provide supplementary details on our training procedures, the network architectures of the Swin Transformer V2 attention module and the ChARM module, as well as an analysis of entropy and the latent representations.

\subsection{Training Details}

The training images are collected from the CLIC dataset \cite{CLIC} and the LIU4K dataset \cite{LIU_dataset}. All training images are rescaled to a resolution of $2000 \times 2000$. We also utilize data augmentation techniques such as rotation and scaling to obtain 81,650 training images with a resolution of $384 \times 384$.

Both Mean Squared Error (MSE) and Multi-Scale Structural Similarity (MS-SSIM) are considered as distortion measures to optimize our models. For MSE optimization, $\lambda$ is chosen from the set ${0.0016, 0.0032, 0.0075, 0.015, 0.03, 0.045, 0.06}$. Each $\lambda$ value corresponds to an independent model targeting a specific bit rate. The number of filters $M$ in the latent representation is set to 160 for the student network and 400 for the teacher network. The number of filters $N$ in the encoder and decoder is set to 128.

For each stage, the models are trained as follows:

Stage 1 (Teacher Network Training): The teacher network is trained for $1.8 \times 10^{5}$ iterations using the loss function in Eq. \ref{teacher_loss}. The learning rate is set to $1 \times 10^{-4}$ and reduced by a factor of 10 at the $1.3 \times 10^{5}$ and $1.6 \times 10^{5}$ iterations.

Stage 2 (Knowledge Distillation Training): The student network is trained jointly with the teacher network for $1.8 \times 10^{5}$ iterations using the loss function in Eq. \ref{total_loss}. The learning rate schedule is the same as in Stage 1. The hyperparameters for the distillation losses are set to $\alpha = 1.0$, $\beta = 0.5$, and $\gamma = 0.5$.

Stage 3 (Student Network Fine-tuning): The student network is fine-tuned independently for an additional $1.5 \times 10^{5}$ iterations without the knowledge distillation loss $L_{\text{KD}}$. The learning rate is further reduced by a factor of 10 at every $5 \times 10^{4}$ iterations.

We use the Adam optimizer with a batch size of 8 throughout all stages. The total training iterations for the student network sum up to approximately $5.1 \times 10^{5}$ iterations to achieve the best results.

The hyperparameters $\alpha$, $\beta$, and $\gamma$ are crucial for balancing the different components of the distillation loss. We empirically found that setting $\alpha = 1.0$ gives sufficient emphasis on matching the reconstructed images, while $\beta = 0.5$ and $\gamma = 0.5$ effectively guide the student network to align its feature maps and latent representations with those of the teacher network.

During the latent space distillation, the selection of the top $C_s$ channels based on entropy ensures that the student network focuses on learning the most informative features. This process leverages the entropy model's ability to quantify the information content in each channel, which is directly related to the channel's contribution to the overall bit rate. By transferring these high-entropy channels, we enable the student network to capture essential information while operating with reduced capacity.

\section{Details on the Swin Transformer V2 Attention Module}
\label{sec_SW2V2}
To effectively capture both local and global dependencies in image data, we integrate the Swin Transformer V2 attention module~\cite{swinTransformerV2} into the transformation layers of our teacher network. Swin Transformer V2 addresses scalability and efficiency limitations of previous transformer models through innovations such as shifted window attention and scaled cosine attention.
In this section, we provide details on its architecture, implementation, and present an ablation study to demonstrate its impact on performance.

\subsection{Architecture Details}

Figure~\ref{fig} illustrates the detailed structures of the Swin Transformer V2 block and its attention module used in our network.

\paragraph{Shifted Window Attention.} The module partitions the feature map into non-overlapping windows and computes self-attention within each window. By shifting the window partition between layers, it enables cross-window connections without significantly increasing computational complexity, balancing local and global context modeling.
\paragraph{Scaled Cosine Attention.} To improve training stability, Swin Transformer V2 replaces the traditional dot-product attention with scaled cosine attention. The attention weights are computed using the cosine similarity between queries and keys, scaled by a learnable temperature parameter $\tau$. The attention output is defined as:

\begin{equation} \label{eq_attention} \text{Attention}(Q, K, V) = \text{Softmax}\left( \frac{\cos(Q, K)}{\tau} + B \right) V, \end{equation}

where $Q$, $K$, and $V$ are the query, key, and value matrices, and $B$ represents the relative position bias.

By integrating the Swin Transformer V2 attention modules after the first and second down-sampling layers in our teacher network, we enhance its capacity to model complex dependencies, leading to improved rate-distortion performance.

\begin{figure}
  \begin{subfigure}{1\linewidth}
    \centering
    \includegraphics[scale=0.35]{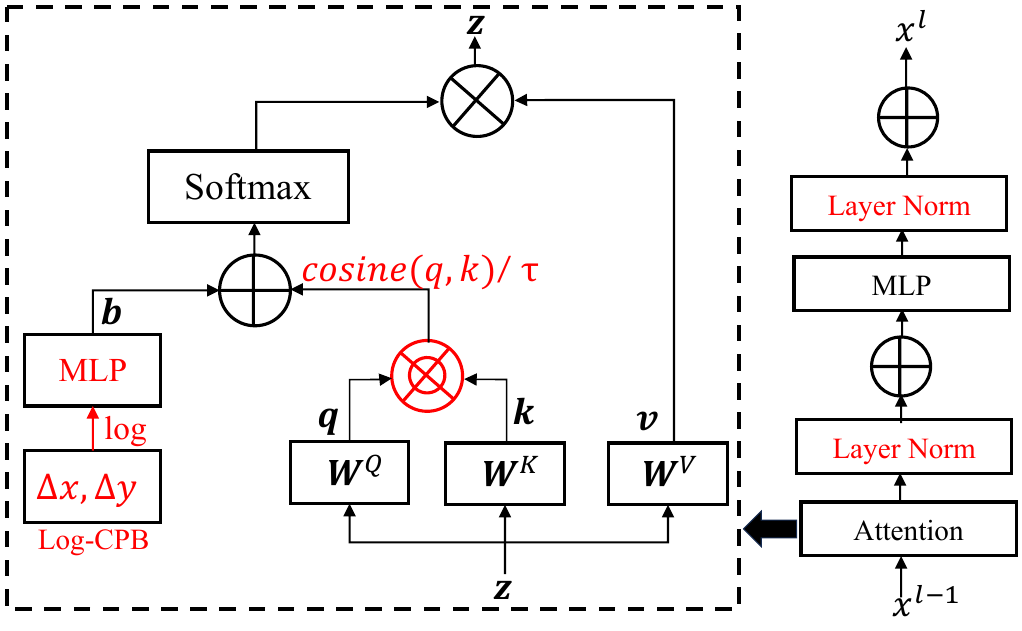}
    \caption{}
    \label{Swin-transformer V2 block}
  \end{subfigure}
  \begin{subfigure}{1\linewidth}
    \centering
    \includegraphics[scale=0.35]{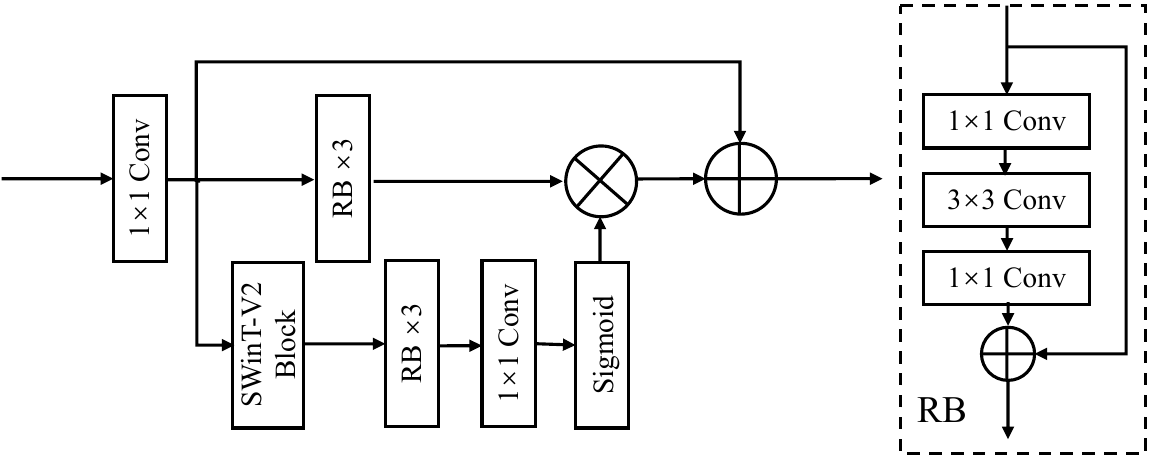}
    \caption{}
    \label{Swin-transformer V2 attention module}
  \end{subfigure}

\caption{The detailed structure of Swin-transformer V2 block and Swin-transformer V2 attention module.}
\label{swinTransformerV2}
\end{figure}

\subsection{Integration into the Teacher Network}

We incorporate the Swin Transformer V2 attention modules after the first and second down-sampling layers in our teacher network. This integration allows the model to capture complex dependencies at multiple scales, enhancing its ability to model intricate patterns in image data.

\subsection{Ablation Study on Attention Modules}

To evaluate the effectiveness of the Swin Transformer V2 attention module in our teacher network, we conducted an ablation study at both low and high bitrates using the Kodak dataset. MSE is used as our optimization loss function. We compared three configurations:

\begin{enumerate} \item \textbf{Baseline}: Teacher network without any attention modules. \item \textbf{Swin Transformer Attention}: Incorporating the original Swin Transformer attention modules~\cite{swinTransformer}. \item \textbf{Swin Transformer V2 Attention}: Incorporating the Swin Transformer V2 attention modules~\cite{swinTransformerV2}. \end{enumerate}

Table~\ref{different_attention_model} summarizes the rate-distortion performance of each configuration at low and high bitrates.

\begin{table}[!hp] 
\centering 
\caption{Ablation study of different attention modules on the teacher network's performance on Kodak dataset.} 
\label{different_attention_model} 
\begin{tabular}{cccc} 
\toprule Module & bpp & PSNR & MS-SSIM  \\ 
\midrule Baseline & 0.199 & 30.65 dB& 13.529 dB   \\
Swin Transformer & 0.201 &  30.71 dB & 13.534 dB \\
Swin Transformer V2 & \textbf{0.198} & \textbf{30.72 dB} & \textbf{13.540 dB} \\ \bottomrule 
Baseline & 0.902 & 30.72 dB & 20.5290 dB  \\
Swin Transformer & 0.903 & 30.75 dB & 20.5345 dB   \\
Swin Transformer V2 & \textbf{0.900} & \textbf{30.81 dB} & \textbf{20.5378 dB}  \\ \bottomrule 
\end{tabular} 
\end{table}

We can draw the following conclusions. At a low bitrate, the Swin Transformer V2 achieves a PSNR improvement of 0.07 dB over the baseline and 0.01 dB over the original Swin Transformer, while also obtaining higher MS-SSIM values. Similarly, at a high bitrate, the Swin Transformer V2 shows a PSNR gain of 0.09 dB over the baseline and 0.06 dB over the original Swin Transformer. It is also noteworthy that our bitrate (bpp) is lower than that of the other two methods at both low and high bitrates.

These improvements demonstrate the superior effectiveness of the Swin Transformer V2 attention module in capturing both local and global contexts, which is crucial for high-quality image reconstruction in compression tasks.





\begin{figure*}[!thp] 
\centering 
\includegraphics[scale=0.4]{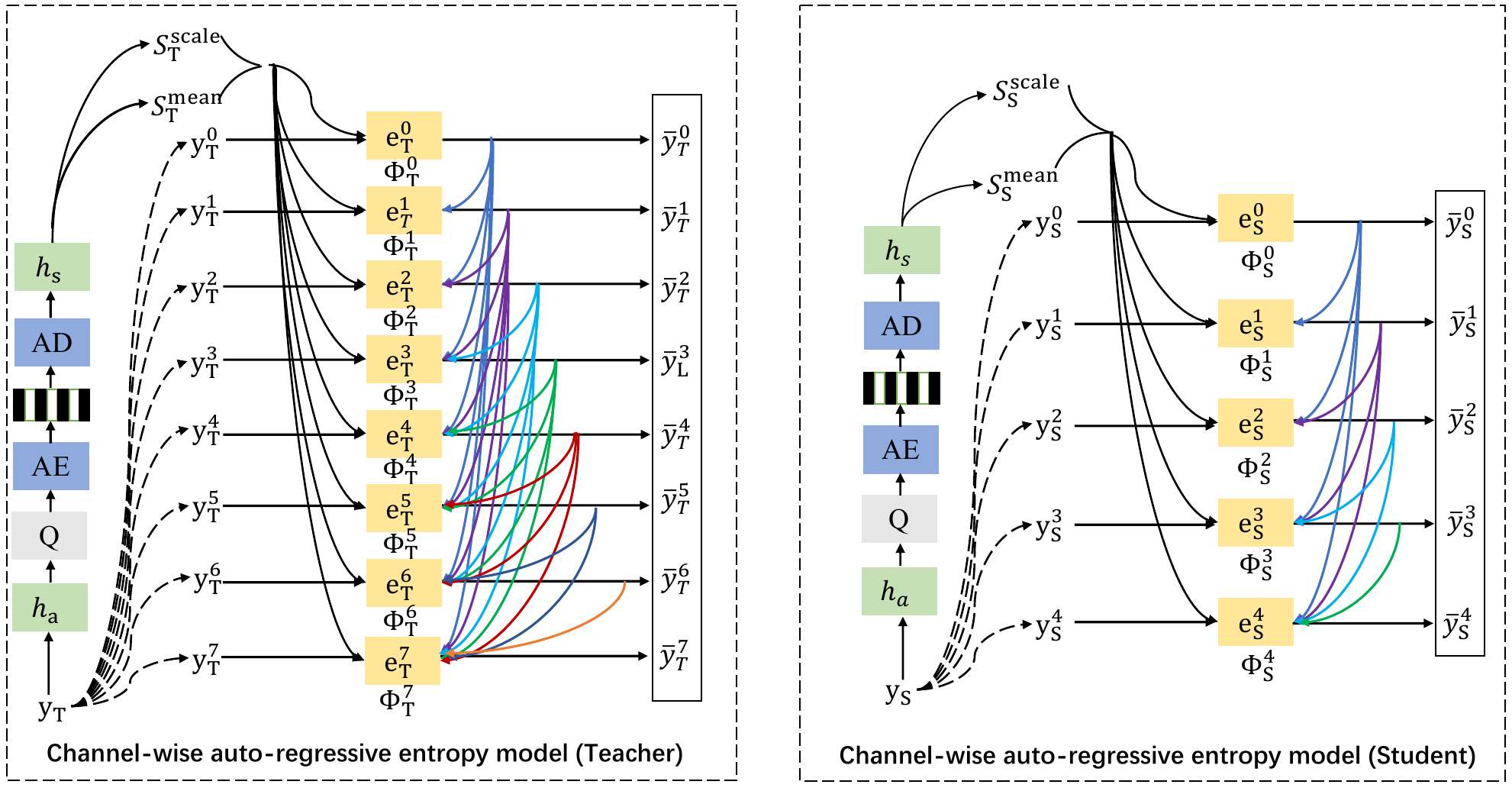} 
\caption{Details of the proposed WeChARM modules for LF and HF subbands.} 
\label{channel_wise_entropy_model} 
\end{figure*}

\begin{figure*}[!thp] 
\centering 
\includegraphics[scale=0.4]{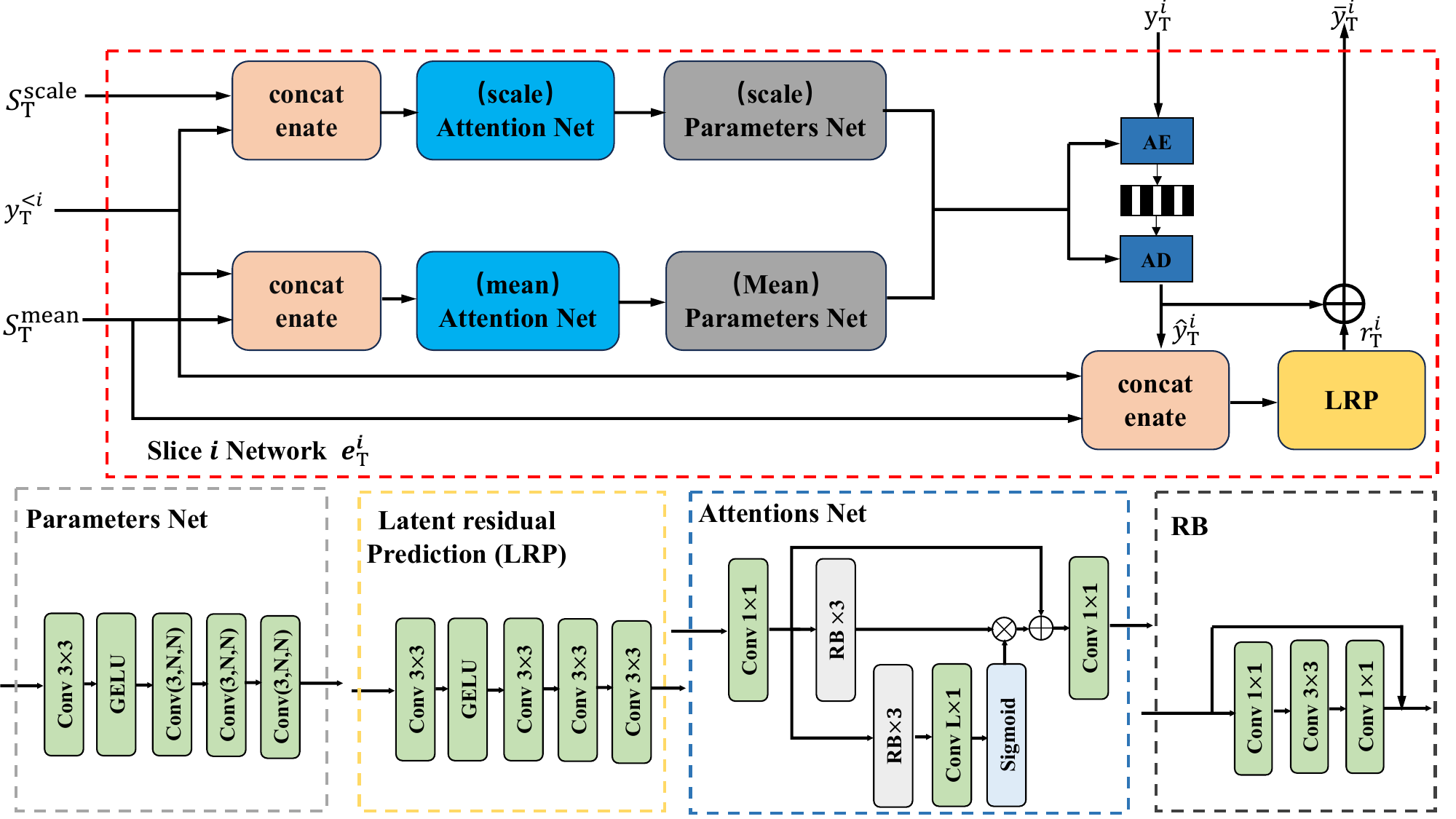} 
\caption{Details of the proposed WeChARM modules for LF and HF subbands.} 
\label{channel_wise_entropy_model_detailed_structure} 
\end{figure*}

\section{ChARM}
\label{ChARM}

In this section, we provide detailed descriptions of the Channel-wise Auto-Regressive Entropy (ChARM) modules used for encoding and decoding the latent representations of the teacher and student models, denoted as $y_{T}$ and $y_{S}$, respectively, as depicted in Figure~\ref{networkstructure} and illustrated in Figure~\ref{channel_wise_entropy_model} and \ref{channel_wise_entropy_model_detailed_structure}.

The ChARM model was first introduced in \cite{channel}, where it employs an auto-regressive strategy for entropy coding by modeling the probability distribution of each channel conditioned on the previously encoded channels. The work in \cite{Liu_2023_CVPR} improved upon this model by integrating a Swin Transformer-based attention mechanism (SWAtten) and reducing the number of slices from 10 to 5, thus achieving a better balance between computational speed and rate-distortion (R-D) performance.

In our approach, we adopt an 8-slice ChARM model to encode $y_{T}$ and a 5-slice ChARM model to encode $y_{S}$, as shown in Figure~\ref{channel_wise_entropy_model}. For the teacher model, each slice consists of 50 channels, while for the student model, each slice consists of 64 channels. We note that the computationally intensive SWAtten module\cite{Liu_2023_CVPR} or the proposed Swin Transformer V2 attention module can be omitted without compromising the R-D performance.

Since the latent representations in the student and teacher networks follow the same encoding logic and differ only in the number of slices and channels per slice, we focus on the encoding and decoding process of the latent representations in the teacher network as an example.

The latent representation $y_{T}$ is divided into slices $y_T^i$ ($i = 0, \dots, 7$), which are sequentially encoded by slice coding networks $e_T^i$. Each slice $y_T^i$ is encoded using side information $S_T^{\text{scale}}$ and $S_T^{\text{mean}}$ obtained from the hyperprior network, assuming that $y_T^i$ follows a Gaussian distribution. Additionally, the encoding process leverages the outputs from preceding slices to reduce inter-slice redundancy.

Figure~\ref{channel_wise_entropy_model_detailed_structure} illustrates the detailed architecture of the slice coding network $e_T^i$. This network is designed to effectively capture the dependencies between slices and model the conditional distributions necessary for accurate entropy coding.


\section{Entropy Analysis of Latent Representations}

In our knowledge distillation framework (illustrated in Figure~\ref{distillation_framework_model}), an essential aspect is the efficient transfer of the most informative features from the teacher network to the student network. To achieve this, we introduce an \textbf{entropy-based distillation loss} that emphasizes the informative channels in the latent representations. In this section, we describe the computation of entropy values for each channel, the process of sorting them in descending order, and the visualization of entropy maps to provide insights into the distribution of information content across different spatial locations and channels.

\begin{figure*}[h]
  \begin{subfigure}{1\linewidth}
    \centering
    \includegraphics[scale=0.3]{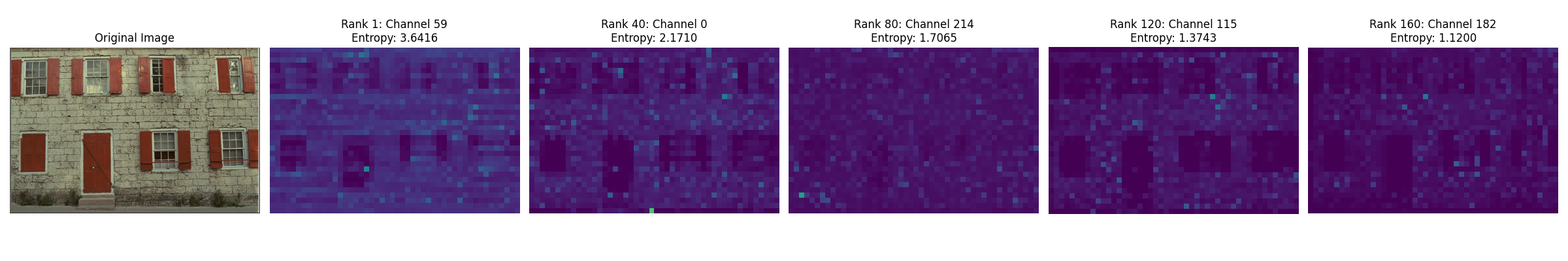}
    \caption{}
    \label{Swin-transformer V2 block}
  \end{subfigure}
  \begin{subfigure}{1\linewidth}
    \centering
    \includegraphics[scale=0.3]{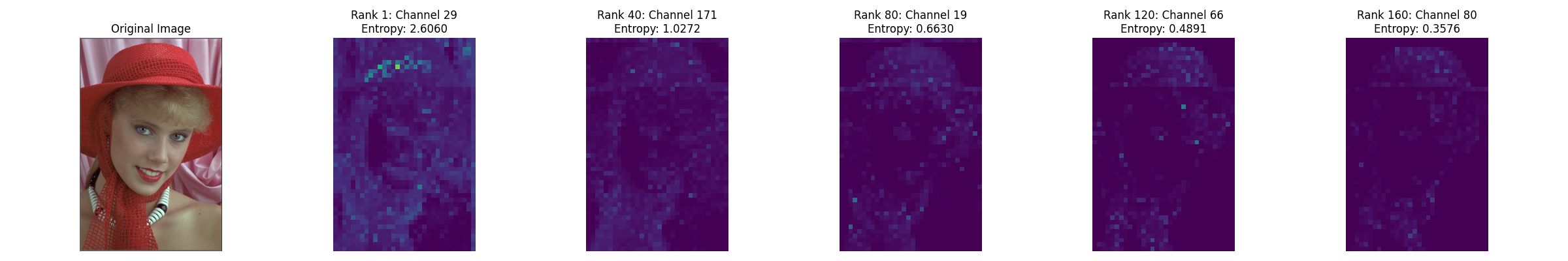}
    \caption{}
    \label{Swin-transformer V2 attention module}
  \end{subfigure}
\begin{subfigure}{1\linewidth}
    \centering
    \includegraphics[scale=0.3]{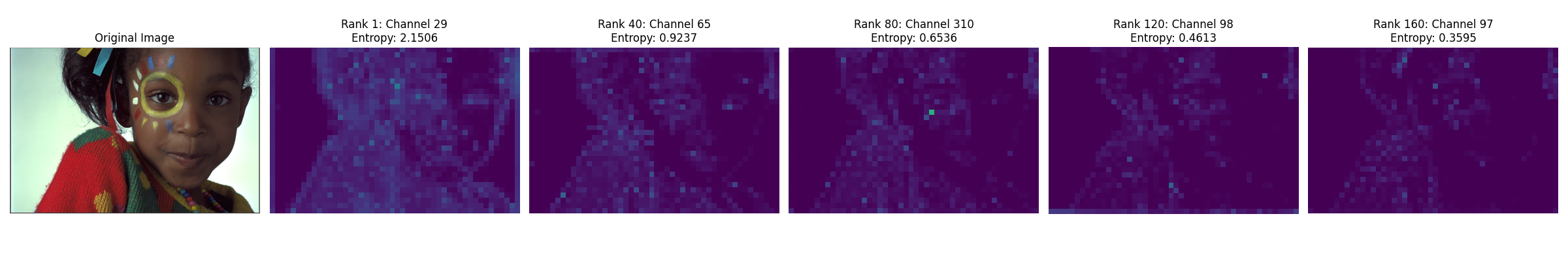}
    \caption{}
    \label{Swin-transformer V2 attention module}
  \end{subfigure}
    \begin{subfigure}{1\linewidth}
    \centering
    \includegraphics[scale=0.3]{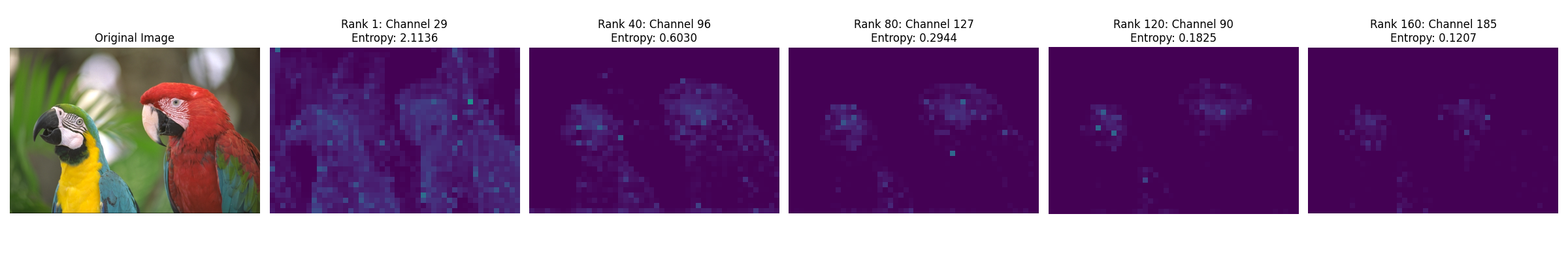}
    \caption{}
    \label{Swin-transformer V2 attention module}
  \end{subfigure}

\caption{Visualization of Selected Channels Based on Entropy Ranking in the Kodak Dataset.}
\label{different_entropy_visualation}
\end{figure*}

\subsection{Entropy Computation}

Given the latent representation $\mathbf{y}$, the mean $\boldsymbol{\mu}$, and the scale $\boldsymbol{\sigma}$ predicted by the hyperprior and context models, we compute the probability density function (PDF) of $\mathbf{y}$ using the Gaussian conditional model:
\begin{equation}
    p_{\mathbf{y}|\boldsymbol{\mu}, \boldsymbol{\sigma}}(\mathbf{y}) = \prod_{i} \left( \Phi\left( \frac{\mathbf{y}_i - \boldsymbol{\mu}_i + 0.5}{\boldsymbol{\sigma}_i} \right) - \Phi\left( \frac{\mathbf{y}_i - \boldsymbol{\mu}_i - 0.5}{\boldsymbol{\sigma}_i} \right) \right),
\end{equation}
where $\Phi(\cdot)$ is the cumulative distribution function (CDF) of the standard normal distribution.

The negative logarithm of the likelihood gives us the entropy map $\mathbf{E}$:
\begin{equation}
    \mathbf{E} = -\log_2 \left( p_{\mathbf{y}|\boldsymbol{\mu}, \boldsymbol{\sigma}}(\mathbf{y}) + \epsilon \right),
\end{equation}
where $\epsilon$ is a small constant added for numerical stability.

\subsection{Channel-wise Entropy Sorting}

For each channel $c$ in the latent representation, we compute the average entropy over all spatial positions:
\begin{equation}
    \bar{E}_c = \frac{1}{H \times W} \sum_{h=1}^{H} \sum_{w=1}^{W} \mathbf{E}_{c,h,w},
\end{equation}
where $H$ and $W$ are the height and width of the entropy map.

We then sort the channels based on their average entropy values in descending order. This ranking allows us to identify the channels that carry the most information, which is valuable for tasks such as channel pruning or importance weighting.

\subsection{Entropy Map Visualization}

To visualize the spatial distribution of entropy, we generate heatmaps for the entropy maps. Specifically, for each channel, we create a heatmap representing the entropy at each spatial location. High entropy regions indicate areas where the latent representation is less certain, often corresponding to complex textures or edges in the image.

In this paper, we randomly selected four images from the Kodak dataset for visualization. For each selected image, we first ranked all channels based on their entropy values in descending order. Subsequently, we chose the channels ranked 1st, 40th, 80th, 120th, and 160th for visualization. The results are shown in Figure \ref{different_entropy_visualation}.

By analyzing the entropy maps and the sorted channel entropies, we gain the following insights:

\begin{itemize}
    \item \textbf{Channel Importance}: Channels with higher average entropy contribute more to the overall information content and are critical for reconstructing high-fidelity images.
    \item \textbf{Spatial Variability}: Regions with high entropy often correspond to complex image areas, such as edges or textures, indicating that the model allocates more bits to encode these regions.
    \item \textbf{Knowledge Distillation Efficiency}: It enables the effective use of distillation strategies to transfer essential information from the Teacher network to the Student network, thereby enhancing the performance of the Student network.
\end{itemize}

\subsection{Implementation Details}

We implemented the entropy computation and visualization using PyTorch and Matplotlib. During the forward pass of the model, we extracted $\mathbf{y}$, $\boldsymbol{\mu}$, and $\boldsymbol{\sigma}$ and computed the entropy maps as described. All channel entropy maps were saved as heatmaps for further analysis. To ensure numerical stability, we added a small constant $\epsilon = 1e^{-9}$ when computing the logarithm.

\subsection{Information-Theoretic Perspective on Our Distillation Strategy}
\label{sec:info_theory}

In our proposed knowledge distillation framework, both feature alignment and entropy-based channel selection can be interpreted through an information-theoretic lens. Specifically, we seek to preserve the critical information content learned by the teacher network and transfer it to the student network under rate constraints. Below, we formally justify why the entropy-based approach and the feature distillation steps are well-motivated in terms of information theory.

\paragraph{Mutual Information and Rate-Distortion.}
Let $\mathbf{X}$ denote the original image, and let $\mathbf{Y}^T$ and $\mathbf{Y}^S$ represent the latent representations of the teacher and student networks, respectively. The teacher network is trained to minimize a rate-distortion objective:
\begin{equation}
    \min_{p(\mathbf{Y}^T|\mathbf{X})} \;\; D(\mathbf{X}, \hat{\mathbf{X}}^T) \;+\; \lambda \, R(\mathbf{Y}^T),
    \label{eq:info_teacher_rd}
\end{equation}
where $D(\cdot,\cdot)$ is a distortion measure, $R(\mathbf{Y}^T)$ is the expected coding cost (in bits), and $\lambda$ is a Lagrange multiplier. From an information theory perspective, this formulation can be viewed as approximating the \emph{rate-distortion function} of the source $\mathbf{X}$, which bounds how accurately one can reconstruct $\mathbf{X}$ given a limited rate budget.

The teacher network’s latent $\mathbf{Y}^T$ thus captures the most relevant information about $\mathbf{X}$ subject to the chosen distortion measure. In other words, it maximizes mutual information $\mathrm{I}(\mathbf{X}; \mathbf{Y}^T)$ under the rate budget:
\begin{equation}
    \mathrm{I}(\mathbf{X}; \mathbf{Y}^T) \;=\; H(\mathbf{Y}^T) \;-\; H(\mathbf{Y}^T \mid \mathbf{X}),
\end{equation}
where $H(\cdot)$ denotes the Shannon entropy and $H(\cdot|\cdot)$ is the conditional entropy. Intuitively, $\mathbf{Y}^T$ is an information-rich, compact representation that is highly predictive of the original image.

\paragraph{Entropy-Based Channel Selection.}
Our entropy-based distillation loss leverages the fact that channels with higher entropies typically encode more salient or complex structures of the image. Let $H_c$ be the entropy of channel $c$ in $\mathbf{Y}^T$, computed as:
\begin{equation}
    H_c \;=\; -\, \mathbb{E}\Bigl[\log_2 \, p\bigl(\hat{y}^T_c \,\bigl\vert\, \hat{z}^T\bigr)\Bigr],
\end{equation}
where $p\bigl(\hat{y}^T_c \mid \hat{z}^T\bigr)$ is the conditional probability model from the teacher’s entropy coding structure, and $\hat{z}^T$ denotes the hyperprior. By ranking all $C_t$ channels in descending order of their entropies $\{H_c\}$, we identify those that carry the largest share of uncertainty—and thus require more bits to encode. These channels often correspond to complex textures, high-frequency details, or critical structural patterns in the image.

When we select the top $C_s$ channels for the student’s latent $\mathbf{Y}^S$, we are effectively focusing on transferring the highest-information subset of the teacher’s representation. This ensures that the student model preserves a near-maximal amount of critical information under a more constrained architecture, improving rate-distortion performance even with fewer overall parameters and channels.

\paragraph{Feature Distillation as KL Minimization.}
Beyond selecting high-entropy channels, our feature-level alignment can be cast as minimizing the Kullback–Leibler (KL) divergence between feature distributions of teacher and student networks:
\begin{equation}
    L_{\text{feature}} \;\propto\; \mathrm{KL}\bigl(f^T_i \;\bigl\|\; f^S_i\bigr),
\end{equation}
where $f^T_i$ and $f^S_i$ denote teacher and student feature maps at layer $i$. Minimizing this KL term encourages the student to replicate the internal feature representation statistics that the teacher found most effective for capturing image structure. Such a principle is consistent with maximizing the mutual information between $\mathbf{X}$ and intermediate network features subject to a complexity budget.

\paragraph{Overall Theoretical Justification.}
By combining:

\begin{itemize}
    \item \textbf{Rate-distortion optimization in the teacher:} ensures $\mathbf{Y}^T$ captures essential information about $\mathbf{X}$ within a given bit budget.
    \item \textbf{Entropy-based channel selection:} targets the most “informative” and high-entropy channels for transfer, aligning with the principle that these channels carry a major portion of the variability and thus code the most critical details of the signal.
    \item \textbf{Feature-level KL minimization:} enforces that the student features remain close to the teacher’s, retaining the most important feature mappings and distributions.
\end{itemize}

From an information-theoretic viewpoint, these components work in tandem to maximize the relevant information preserved in the student latent representations, effectively closing the performance gap between the high-capacity teacher and the lightweight student. Empirical results (Sec.~\ref{sec_results}) confirm that this strategy indeed yields a minimal BD-Rate penalty while substantially reducing the model size and computational overhead.

\end{document}